\documentclass{article}
\usepackage{spconf,amsmath,epsfig}

\begin{document}\sloppy

% Example definitions.
% --------------------
\def\x{{\mathbf x}}
\def\L{{\cal L}}

% Title.
% ------
\title{Joint Haze Image Synthesis and Dehazing with MMD-VAE Losses$^\star$}
%
% Single address.
% ---------------
\name{Zongliang Li$^{\dagger}$, Chi Zhang$^{\dagger}$, Gaofeng Meng$^{\ddagger}$, Yuehu Liu$^{\dagger}$
\thanks{$^\star$This work is supported by National Nature Science Foundation of China (Grant No: 91520301). Chi Zhang is  the corresponding author. {\tt\small chi.zhang.cn@ieee.org}}% <-this % stops a space
 }
\address{$^{\dagger}$Institute of Artificial Intelligence and Robotics, Xi'an Jiaotong University, Xi'an, China   \\ $^{\ddagger}$National Laboratory of Pattern Recognition, Institute of Automation,\\ Chinese Academy of Sciences, Beijing, China}
\maketitle

\begin{abstract}
Fog and haze are weathers with low visibility which are adversarial to the driving safety of intelligent vehicles equipped with optical sensors like cameras and LiDARs.
Therefore image dehazing for perception enhancement and haze image synthesis for testing perception abilities are equivalently important in the development of such autonomous driving systems. 
From the view of image translation, these two problems are essentially dual with each other, which have the potentiality to be solved jointly. 
In this paper, we propose an unsupervised Image-to-Image Translation framework based on Variational Autoencoders (VAE) and 
Generative Adversarial Nets (GAN) to handle haze image synthesis and haze removal simultaneously. 
Since the KL divergence in the VAE objectives could not guarantee the optimal mapping under imbalanced and unpaired training samples with limited size, 
Maximum mean discrepancy (MMD) based VAE is utilized to ensure the translating consistency in both directions. 
The comprehensive analysis on both synthesis and dehazing performance of our method demonstrate the feasibility and practicability of the proposed method.
\end{abstract}

\begin{keywords}
	Image-to-Image Translation, MMD-VAE, Cycle-consistency, Haze Image Synthesis, Image Dehazing
\end{keywords}
\section{Introduction}
\label{sec:intro}
Foggy and hazy weather brings huge troubles to the driving safety of vehicles, especially for automated vehicles equipped with color cameras and LiDARs. 
The dense-distributed atmospheric suspended particulates may cause low visibility and sensory noises, which results in degraded sensory data.

To ensure the robust visual perception in such extreme situations, haze removal is a prerequisite for image enhancement. 
Motivated by haze image formation mechanism, many sophisticated handy-crafted algorithms\cite{He2009Single, Meng2014Efficient} and 
learning-based methods\cite{Cai2016DehazeNet, li2018single} have been proposed for haze removal. 
On the other hand, it is also necessary to validate perceptual algorithms/modules with haze images. 
However, the acquisition of sufficient, diverse image data under such scenarios is difficult since the foggy and hazy weather is rare, temporary and hard to forecast. 
Instead, a common practice is to synthesize haze images from clear images and corresponding depth maps under haze image formation model, 
with respect to certain atmospheric scattering coefficient and airlight intensity. 

In practice, the result images in both tasks draw more attention than other byproducts like transmission map and airlight intensity. 
To this end, we can simplify these two problem as particular Image-to-Image Translation problem which translate image between haze image domain and haze-free image domain. 
As a pair of corresponding haze image and clear image represent the same scene, we assume they can be mapped to a same latent representation in share-latent space\cite{Liu2017Unsupervised}. 
Based on this assumption, we propose a method based on VAEs and GANs to joint solve these two problems. 
In our method, we use two VAE-GAN to model haze image and haze-free image and the adversarial training objective enforces the generated image faithful for corresponding image domain and introduce the MMD-VAE cycle-consistency loss to further regularize the model. 

The acquisition of desirable and diverse haze images in the traffic environment is of great difficulties due to the speciality, transience of such rare weather. 
Moreover, the corresponding pairs of a clear image and a haze image could not be obtained because of the spatio and temporal changes. 
In this case, the traditional VAE may overestimate the variance of sample data, given such unpaired and size-limited training set. 
Instead of the KL divergence, the Maximum mean discrepancy (MMD) based VAE losses are utilized in our framework, which overcome the problem and be able to learn informative latent code.

The contribution of this work are as follows:
\begin{itemize}
	\item We proposed an end-to-end method based on VAEs and GANs to jointly solve the haze image synthesis problem and haze removal problem with the condition of lack of hazy traffic images.
	\item To overcome the problems result from dataset of limited size, we introduce MMD-VAE losses in the image translation process.
\end{itemize}

\begin{figure*}[!tbh]%[!htp]
	\centering
	\includegraphics[width=0.7\textwidth]{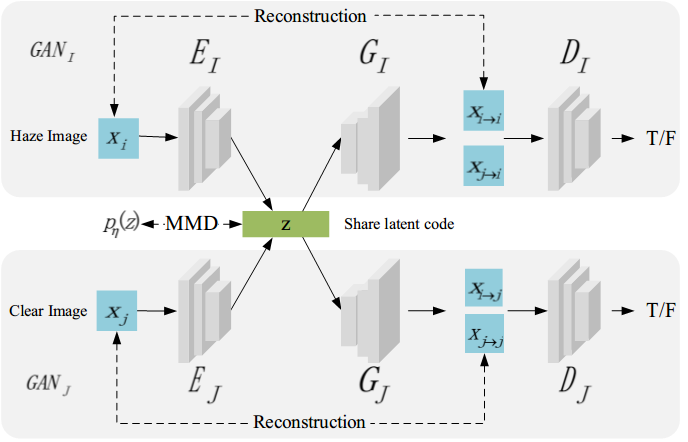}
	\caption{Our framework. Given any pair of images of haze image domain $\mathcal{X}_I$ and clear image domain $\mathcal{X}_J$. 
	The encoders $E_I$ and $E_J$ map the images to the share-latent space from where the generators $G_I$ and $G_J$ reconstructed the input image, respectively. 
	$D_I$ and $D_J$ are adversarial discriminators for the corresponding domain, in charge of evaluating if the translated image are realistic.}
	\label{fig:framework}
\end{figure*}

\section{Related work}
\subsection{Haze Image Formation Model}
In computer vision and computer graphics, the model widely used to describe the formation of haze image\cite{He2009Single} is as follows:
\begin{equation}
	I(x) = J(x)t(x) + A(1-t(x))
	\label{equ:hazemodel}
\end{equation}
where $I(x)$ and $J(x)$ represent the haze image and the scene radiance (i.e. the original clear image), respectively. 
$A$ is the global atmospheric light, and $t(x)$ is the medium transmission map. If the haze is homogeneous, 
the transmission map can be expressed as $t(x)=e^{-\beta d(x)}$, where $\beta $ is the scattering coefficient and $d(x)$ is the scene depth. 
As only the observed image $I(x)$ is known, recovering the scene radiance $J(x)$ is highly ill-posed.

Based on this model, some sophisticated hand-crafted methods like \cite{He2009Single} are proposed which get a significant advances in image dehazing. 
These method all use hand-crafted features, which are mainly based on chromatic, textural and contrast properties, to estimate transmission maps and atmospheric light. 
Meng \cite{Meng2014Efficient} propose a method for single image dehazing benefiting from the exploration on the inherent boundary constraint on the transmission function. 
He et al.\cite{He2009Single} present an interesting image prior - \textit{dark channel prior} for single image dehazing, achieve a quite compelling haze-free result of very high quality. 
However, as the assumption on hand-crafted features do not always hold, these methods do not always works. 
For example, the method proposed by He et al.\cite{He2009Single} does not work well for the object which are similar to the airlight. 

\subsection{Image to Image Translation}
Proposed by Isola et al.\cite{Isola2016Image} firstly, Image to Image Translation is the learning problem which maps an image in one domain to a corresponding image in another domain. 
The problem can be studied in supervised and unsupervised learning settings. In the supervised settings, paired of corresponding images in different domains are requirement\cite{Ledig2017Photo,Isola2016Image}. 
The “pix2pix” framework of Isola et al.\cite{Isola2016Image} uses supervised settings and conditional generative adversarial network to learn the mapping from input images to output images. 
In the unsupervised setting, we only need two independent sets of images from two domains respectively. The unsupervised Image-to-image translation problem is more applicable since training data is easier to acquire. 
This problem is inherently ill-posed, and additional constrains are necessary to solve it. A popular constraint is the cycle consistency loss\cite{Zhu2017Unpaired,Liu2017Unsupervised,Yi2017DualGAN}, assumed in CycleGAN\cite{Zhu2017Unpaired} and Johnson’s\cite{Johnson2016Perceptual} work. 
As for lack of paired images in traffic scene, we choose unsupervised setting to handle our problem.

\section{Proposed Method}
There exists no one-to-one mapping between images in clear domain $\mathcal{X}_J$ and haze domain $\mathcal{X}_I$ with unsupervised setting of image translation task. 
In this case, without reasonable assumption, the exact mapping between clear images and haze images could not be learned properly. 
By posing a latent space of lower dimension (e.g. 8-dimension vector space) in the image translation process, through which the translation between different domains or backward to the same domain should be passed, 
the space of the domain-invariant content is greatly reduced. Based on such shared-latent space assumption in\cite{Liu2017Unsupervised}, 
in this paper we introduce a latent space $\mathcal{Z}$ as the middle domain between the clear image domain $\mathcal{X}_J$ and the haze image domain $\mathcal{X}_I$.

The framework of our method is shown in Figure~\ref{fig:framework}. Given any pair of haze image $x_i \in \mathcal{X}_I$ and clear image $x_j \in \mathcal{X}_J$, 
the encoders $E_I$ and $E_J$ aim to map the images to a shared latent code $z$ in a share-latent space $\mathcal{Z}$ as $z =E_I(x_i) =E_J(x_j)$, meanwhile the generators $G_I$ and $G_J$ can recover both images from the latent code: $x_i =G_I(z), x_j = G_J(z)$. 
In this model, the haze image synthesis function $x_i=F_{J\rightarrow I}(x_j)$ that map from clear image to haze image can be represented by $F_{J\rightarrow I}(x_j)=G_I(D_J(x_j))$. 
Similarly, haze removal function can be represented by $F_{I\rightarrow J}(x_i)=G_J(D_I(x_i))$. 
Equally, in our structure there also are two reconstruct streams for haze image domain $\mathcal{X}_I$ and clear image domain $\mathcal{X}_J $, 1) $x_{i \rightarrow i} = G_I(E_I(x_i))$ and $x_{j \rightarrow j} = G_J(E_J(x_j))$ which are reconstruction stream 
and 2)  $x_{i \rightarrow j \rightarrow i}=F_{J\rightarrow I}(F_{I\rightarrow J}(x_i))$ and $x_{j \rightarrow i \rightarrow j}=F_{I\rightarrow J}(F_{J\rightarrow I}(x_j))$ which are cycle-reconstruction stream. 
The discriminators $D_I$ and $D_J$ are used to distinguish between the generated images and the real images for hazy and clear domain, respectively. 
These subnetworks consist two Generative Adversarial Networks: $\text{GAN}_I: \{E_I, G_I, D_I\}$ for haze image domain, and $\text{GAN}_J: \{E_J, G_J, D_J\}$ for clear image domain. We apply adversarial losses to both GANs as:
\begin{equation}
\begin{aligned}
\mathcal{L}_{\text{GAN}_I}(E_I, G_I, D_I)= \lambda_{adv} E_{x_i\sim P_{X_I}}[log D_I(x_i)] 
\\ + \lambda_{adv} E_{z_j\sim q_J(z_j|x_j)}[log(1-D_I(G_I(z_j)))]
\end{aligned}
\end{equation}
\begin{equation}
\begin{aligned}
\mathcal{L}_{\text{GAN}_J}(E_J, G_J, D_J)= \lambda_{adv} E_{x_j\sim P_{X_J}}[log D_J(x_j)] 
\\ + \lambda_{adv} E_{z_i\sim q_I(z_i|x_i)}[log(1-D_J(G_J(z_i)))]
\end{aligned}
\end{equation}

Generally, the adversarial loss can not guarantee a specific mapping from source domain to target domain\cite{Zhu2017Unpaired, Liu2017Unsupervised}. 
That is to say, given an input sample $x_i$ from source domain, the network only with adversarial loss could be mapped to any sample from the target distribution, rather than a desired output $y_i$. 
To address this problem, a constraint of forward-backward consistency\cite{Zhu2017Unpaired} is introduced to reduce the dimension of the mapping space, inspired by ``back translation and reconciliation'' in the field of human language translation. 
For example, haze image synthesis function $F_{J \rightarrow I}$ and haze removal function $F_{I\rightarrow J}$ are further constrained by cycle-consistency loss: $x_i = F_{J\rightarrow I}(F_{I\rightarrow J}(x_i))$ and $x_j = F_{I\rightarrow J}(F_{J \rightarrow I}(x_j))$. 
In other word,  the input image can be reconstructed from translating back to the translated input image.

\subsection{MMD-VAE Losses}
In our method, the latent code $z$ is learned automatically by a process of variational autoencoder(VAE). 
The encoder-generator pair $\{E_J, G_J\}$ constitute a VAE for clear domain $\mathcal{X}_J$: $\text{VAE}_J$ where the encoder $E_J$ maps $x_j$ to latent code in latent space $z_j$ and the generator $G_J$ reconstruct the input image from random-perturbed version of latent code. 
In our formulation, the distribution of the latent code $z_j$ is a  Gaussian with unit variance, written as $q_{J}(z_j|x_j)\equiv N(z_J|E_{\mu,J}(x_j),I)$ where $I$ is an identity matrix and mean vector $E_{\mu, J}(x_j)$ is produced by the encoder. 
The latent code $z_j$ is sampled via $z_j\sim q_J(z_j|x_j)$. Thus, the reconstructed image is $x_{j\rightarrow j}=G_J(z_j\sim q_J(z_j|x_j))$. 
Similarly,  $\{E_I, G_I\}$ constitute a VAE for hazy domain $\mathcal{X}_I$:$\text{VAE}_I$. The encoder $E_I$ products a mean vector $E_{\mu, I}(x_i)$ and the latent code $z_i$ is sampled by $q_{J}(z_j|x_j)\equiv N(z_J|E_{\mu,J}(x_i),I)$. The reconstructed image is $x_{i\rightarrow i}=G_J(z_i\sim q_I(z_i|x_i))$. 
With the reparameterization trick\cite{kingma2013auto}, the sample operation $z_i\sim q_I(z_i|x_i)$ and $z_j\sim q_J(z_j|x_j)$ can be implemented by $z_i=E_{\mu, I}(x_i) + \eta$ and $z_j=E_{\mu, J}(x_j) + \eta$ where $\eta \sim \mathcal{N}(\eta|0,I)$.

In the case of limited training sample, the evidence lower bound objective (ELBO) of traditional VAE tends to over estimation the variance of the latent feature $z$. 
Moreover, the ELBO encourages both encoder and generator to under use the established latent space $z$. 
Both drawbacks result from the use of KL-divergence when calculate the statistic distance between the posterior distribution and the prior\cite{Zhao2017InfoVAE}. 
To overcome these drawbacks, in this paper, Maximum mean discrepancy VAE (MMD-VAE) is introduced to constrain the forward-backward consistency previously stated.

\subsubsection{MMD-VAE}
MMD is the supremum of the difference between the expectation $Ep\left[f(x)\right]$ and $Ep\left[f(y)\right]$, which is the random projection (via a function $f$) of the source sample $X$ and target sample $Y$. The theoretic definition of MMD is as follows:
\begin{equation}
\begin{aligned}
\text{MMD}[\mathcal{F},p,q] = \text{sup}_{f\in \mathcal{F}}(E_p[f(x)]-E_q[f(y)])\\
\text{MMD}[\mathcal{F},X,Y] = \text{sup}_{f\in \mathcal{F}}(\frac{1}{m}\sum_{i=1}^m f(x_i)-\frac{1}{n} \sum_{i=1}^n f(y_i))
\end{aligned}
\end{equation}
where $\mathcal{F}$ is the set of all mapping functions which map the feature space to real number set $R$. If and only if the distributions of $x$ and $y$ are identical, the MMD equals 0. MMD is optimal when the space $F$ of each projection $f$ is constructed as a unit sphere in Reproducing Kernel Hilbert Space, i.e.
\begin{equation}
\begin{aligned}
f(x) = \left \langle f, \phi(x) \right \rangle_\mathcal{H}
\end{aligned}
\end{equation}
Based on such Kernel embedding of distributions, MMD could be derived as
\begin{equation}
\begin{aligned}
\text{MMD}[\mathcal{F},p,q] =\Vert \mu_p - \mu_q\Vert_\mathcal{H}\\
\end{aligned}
\label{equ:mmd1}
\end{equation}
\begin{equation}
\begin{aligned}
\text{MMD}^2[ \mathcal{F},p,q] = E_p\left\langle\phi(x),\phi(x')\right\rangle_\mathcal{H} \\
+ E_q\left\langle\phi(y),\phi(y')\right\rangle_\mathcal{H} - 2E_{p,q}\left\langle\phi(x),\phi(y)\right\rangle_\mathcal{H}
\end{aligned}
\end{equation}
In practical we select the Guassian Kernel for such high-dimensional RKHS $k(x,x')=\text{exp}(- \frac{\Vert x-x'\Vert^2}{2\sigma^2})$, where the MMD constrain could be calculated as follows,
\begin{equation}
\begin{aligned}
\text{MMD}^2[\mathcal{F},X,Y]=\frac{1}{m(m-1)}\sum_{i\neq j}^m k(x_i, x_j)\\
+ \frac{1}{n(n-1)}\sum_{i\neq j}^m k(y_i, y_j) - \frac{2}{mn}\sum_{i,j=1}^{m,n} k(x_i, y_j)
\end{aligned}
\label{equ:mmd_constraint}
\end{equation}
Thus, the MMD-VAE loss object as:
\begin{equation}
\begin{aligned}
\mathcal{L}_{\text{VAE}_I}(E_I, G_I) = \lambda_m \text{MMD}(q_I(z_i|x_i) \Vert p_{\eta}(z))
\\  - \lambda_{recon} E_{z_i \sim q_I(z_i|x_i)}[\log p_{G_I}(x_i|z_i)]
\end{aligned}
\label{equ:mmd1loss}
\end{equation}
\begin{equation}
\begin{aligned}
\mathcal{L}_{\text{VAE}_J}(E_J, G_J) = \lambda_m \text{MMD}(q_J(z_j|x_j) \Vert p_{\eta}(z))
\\  - \lambda_{recon} E_{z_j \sim q_J(z_j|x_j)}[\log p_{G_J}(x_j|z_j)]
\end{aligned}
\label{equ:mmd2loss}
\end{equation}
Where $\lambda_m$, $\lambda_{recon}$ controls the weight of the loss terms, and the $\text{MMD}$ terms penalized the latent code deviating from the prior distribution. The negative log-likelihood objective term ensure the reconstructed image resembles the input one.

Based on the MMD-VAE, the cycle-consistency loss are defined as:
\begin{equation}
\begin{aligned}
\mathcal{L}_{cc_I}(E_I, G_I, E_J, G_J) = \lambda_m \text{MMD}(q_I(z_i|x_i) \Vert p_{\eta}(z))
\\  + \lambda_m \text{MMD}(q_J(z_j| x_{i \rightarrow j})\Vert p_{\eta}(z))
\\  - \lambda_{recon} E_{z_j \sim q_J(z_j|x_{i \rightarrow j})}[\log p_{G_I}(x_i|z_j)]
\end{aligned}
\label{equ:cc1loss}
\end{equation}
\begin{equation}
\begin{aligned}
\mathcal{L}_{cc_J}(E_J, G_J, E_I, G_I) = \lambda_m \text{MMD}(q_J(z_j|x_j) \Vert p_{\eta}(z))
\\  + \lambda_m \text{MMD}(q_I(z_i| x_{j \rightarrow i})\Vert p_{\eta}(z))
\\  - \lambda_{recon} E_{z_i \sim q_I(z_i|x_{j \rightarrow i})}[\log p_{G_J}(x_j|z_i)]
\end{aligned}
\label{equ:cc2loss}
\end{equation}
Where the $\text{MMD}$ terms penalize the latent code deviating from the prior distribution, and the negative log-likelihood objective term ensure the twice translated image resembles the input one.

\subsection{Total loss}
The full object objective, jointly optimized by the encoders, decoders and discriminators, is a weighted sum of the MMD-VAE loss, the adversarial loss, and the cycle-consistency loss terms.
\begin{equation}
\begin{aligned}
\min_{E_I,E_J,G_I,G_J}\max_{D_I,D_J} \mathcal{L}(E_I, E_J, G_I, G_J, D_I, D_J) = \\
  \mathcal{L}_{\text{VAE}_I}(E_I, G_I) + \mathcal{L}_{\text{VAE}_J}(E_J, G_J) \\
+ \mathcal{L}_{\text{GAN}_I}(E_I, G_I, D_I) + \mathcal{L}_{\text{GAN}_J}(E_J, G_J, D_J)\\
+ \mathcal{L}_{cc_I}(E_I, G_I, E_J, G_J) + \mathcal{L}_{cc_J}(E_J, G_J, E_I, G_I)
\end{aligned}
\label{equ:loss}
\end{equation}

\section{Experiment}
We quantitatively and qualitatively evaluate our method, in comparison with several state-of-the-art algorithms from two aspects, i.e. the synthesis and haze removal performance respectively. 
The metrics used in quantitative evaluation are Peak Signal to Noise Ratio(PSNR) and Structural Similarity Index(SSIM).

\subsection{Implementation Details}
\subsubsection{Dataset}
We use the clear images and correspond depth images in Apollo\cite{apolloscape_arXiv_2018} dataset to synthesis the haze images. 
2000 traffic scene images and corresponding depth images are randomly chosen from Apollo Dataset. Given a clear image $J$ and the corresponding ground truth depth $d$, we synthesize a haze image $I$ according to \eqref{equ:hazemodel}. 
We generate the random atmospheric light $A=[n_1,n_2,n_3]$, where $n\in [0.8,1.0]$, and use random scattering coefficient $\beta \in [0.8,1.6]$ for each image. 
We randomly choose 1500 clear images and the synthesized haze images as training set with rest as test set.
\subsubsection{Training Details}
We trained our network with Adam optimizer where the learning rate is 0.0001 and momentums are set to 0.5 and 0.999. 
The batch size is 1, which means each mini-batch consist of one haze image and one clear image. 
The value for other parameters are $\lambda_m = 0.01$, $\lambda_{adv} = 1$, and $\lambda_{recon} = 10$. The network were trained on NVIDIA TITAN Xp GPU with 12GB GPU memory for 250K iterations and save the model every 10K iterations. 
The best model was chosen as the final model using in the experiment.

\begin{table}
\begin{center}
\caption{Quantitative comparisons on testing set.}
\label{tab:result}
\begin{tabular}{|c|c|c|c|c|}
\hline
 & \multicolumn{2}{|c|}{Hazy Synthesis} & \multicolumn{2}{|c|}{Haze Removal}\\
\cline{2-5}
 & SSIM & PSNR & SSIM & PSNR \\
\hline
UNIT & 0.8524 & 24.5181 & 0.8255 & 21.1384 \\
\hline
CycleGAN & 0.8858 & 25.1917 & 0.8432 & 22.6649\\
\hline
He's & N/A & N/A & 0.8338 & 18.2011\\
\hline
Meng's & N/A & N/A & 0.8234 & 18.4564\\
\hline
DehazeNet & N/A & N/A & \textbf{0.8832} & 20.2258\\
\hline
Ours & \textbf{0.9271} & \textbf{27.3772} & 0.8801 & \textbf{23.6779}\\
\hline
\end{tabular}
\end{center}
\end{table}

\begin{figure*}[!tbh]%[!htp]
	\centering
	\includegraphics[width=\textwidth]{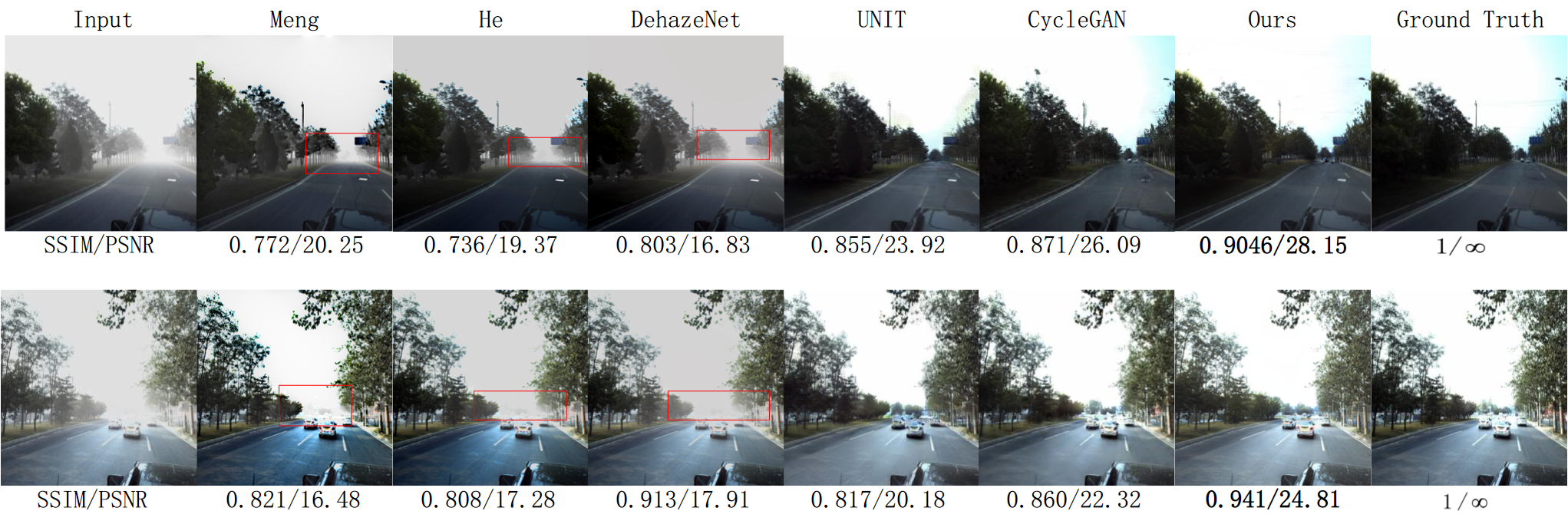}
	\caption{Two images from synthetic hazy traffic scene and the haze removal results of serval state-of-the-art dehazing methods. 
	The detail differences are in rectangles and the quantitative result are shown blow the result. }
	\label{fig:dehaze}
\end{figure*}
\begin{figure*}
	\centering
	\includegraphics[width=\textwidth]{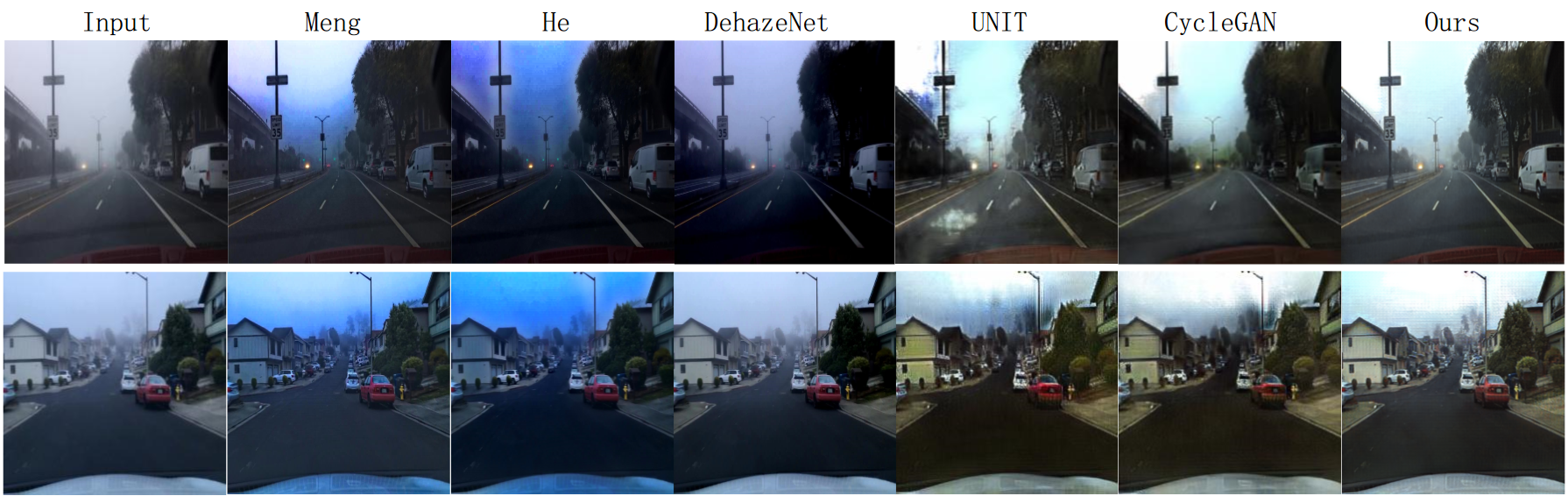}
	\caption{Real world hazy traffic images and corresponding haze removal results of several state-of-the-art methods.}
	\label{fig:dehaze_real}
\end{figure*}

\begin{figure*}[!tbh]%[!htp]
	\centering
	\includegraphics[width=\textwidth]{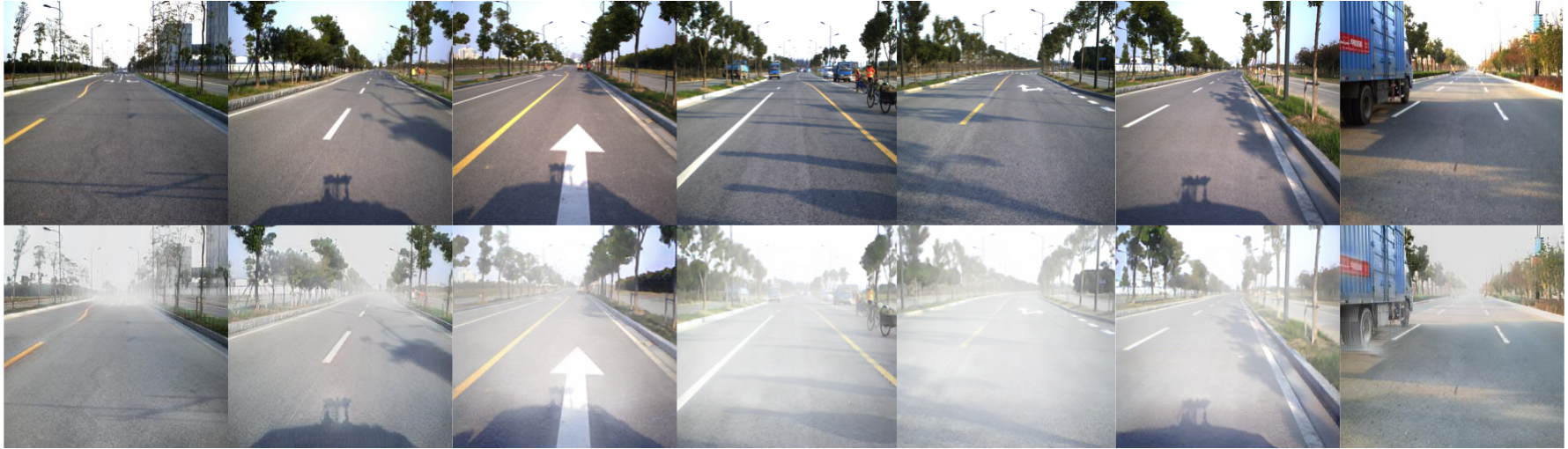}
	\caption{haze images synthesis result of our method on real world traffic scene. The first row is the input images, and the second row is generated haze images.} 
	\label{fig:our_result}
\end{figure*}

\begin{figure}[!tbh]%[!htp]
	\centering
	\includegraphics[width=0.5\textwidth]{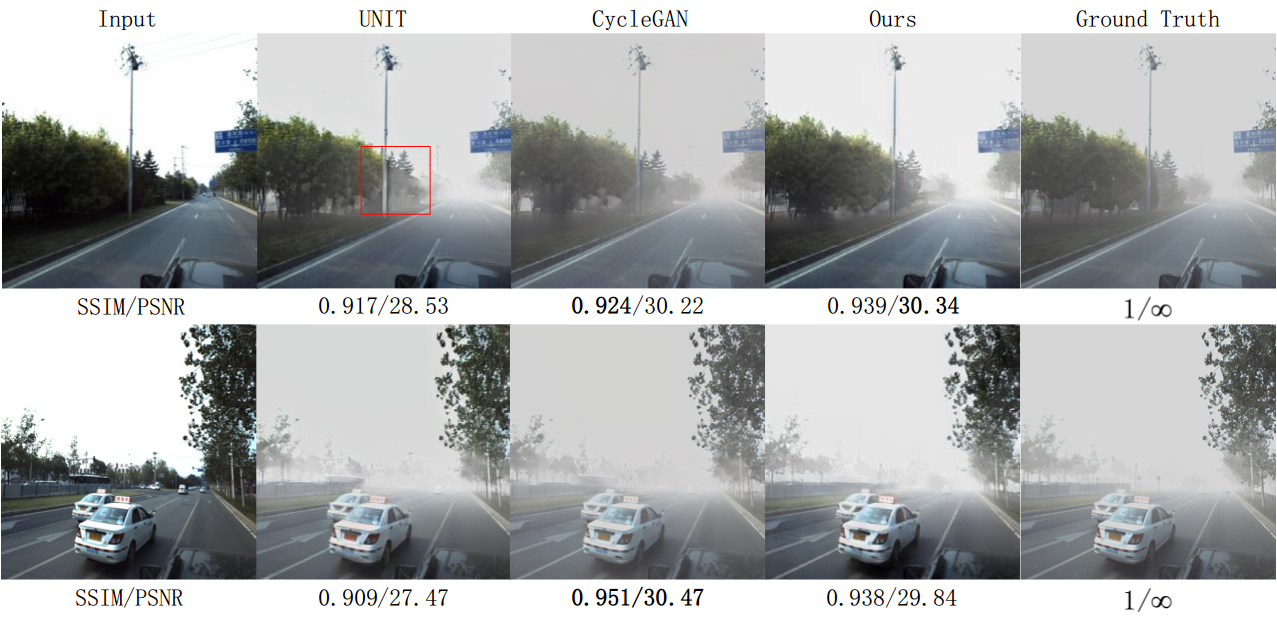}
	\caption{ Quantitatively compare the haze image synthesis result with CycleGAN and UNIT.} 
	\label{fig:synHaze}
\end{figure}

\subsection{Haze Images Synthesis}
We quantitatively and qualitative compare our method with serval state-of-the-art methods, CycleGAN and UNIT, on haze image synthesis using PSNR and SSIM. 
The Quantitatively evaluation results are shown in Table~\ref{tab:result}. Our method generates the haze image with higher PSNR and SSIM score than other methods. 
In Figure~\ref{fig:synHaze}, we show two images from testing set and corresponding generated haze images. 
The result by UNIT\cite{Liu2017Unsupervised} have some artifacts(e.g., the part enclosed in red box on the image of first row). 
The hazy synthesis results by CycleGAN\cite{Zhu2017Unpaired} are more similar to the ground truth at the overall brightness. 
But our method can generated more diversity samples as shown in Figure~\ref{fig:our_result}. 

\subsection{Haze Removal}
We also evaluate the haze removal result of our method and serval dehazing methods, He's method\cite{He2009Single}, Meng's method\cite{Meng2014Efficient} and DehazeNet\cite{Cai2016DehazeNet} quantitatively and qualitatively. 
As shown in Table~\ref{tab:result}, our method achieves higher score than other methods. Two examples of synthetic haze images and corresponding dehazing result are shown in Figure~\ref{fig:dehaze}. 
In results by He\cite{He2009Single} and Meng\cite{Meng2014Efficient}, the color of road surface distort heavily, which is mainly caused by the inaccurate transmission maps. 
DehazeNet\cite{Cai2016DehazeNet} use CNN to estimate transmission maps, which overcomes the fault of\cite{He2009Single, Meng2014Efficient} to some extent. 
Thus, the result of DehazeNet\cite{Cai2016DehazeNet} has few color distortions. But the results still contains some hazy residuals near vanishing point(the area enclosed in red box in Figure~\ref{fig:dehaze}). 
In contrast, few hazy residuals and artifacts appear in the images generated by our method. UNIT\cite{Liu2017Unsupervised} and CycleGAN\cite{Zhu2017Unpaired} get a similar result.

The qualitatively comparisons, which test on the real world haze images, are shown in Figure~\ref{fig:dehaze_real}. 
The same with the results on synthetic haze images, the results of \cite{He2009Single,Meng2014Efficient} have heavily color distortions, 
DehazeNet\cite{Cai2016DehazeNet} can not restore the area near the vanishing point and there are some artifacts in the result of UNIT\cite{Liu2017Unsupervised} and CycleGAN\cite{Zhu2017Unpaired}. 
In contrast, the images generated by our method are much clearer and realistic than the methods above. 

\section{Conclusions}
In this paper, We proposed a model which is constrained by MMD-VAE cycle-consistency, to jointly handle the foggy image synthesis and the haze removal of the traffic scene. 
With this model, we can synthesize enough foggy traffic scene for self-driving car testing and remove haze to improve the performance of the self-driving car. 
In our future work, we will handle more extreme weather scene synthesis problem, such as rainy, snowy and dark, based on the proposed model. 

\bibliographystyle{IEEEbib}
\bibliography{icme2019template}

\end{document}